\documentclass[letterpaper, 10 pt, conference]{ieeeconf}
\IEEEoverridecommandlockouts
\usepackage{amsmath,amsfonts}
\usepackage{amsmath} % for 'bmatrix' environment
\usepackage{newtxtext,newtxmath} % optional -- Times Roman clone
\usepackage{algorithmic}
\usepackage{algorithm}
\usepackage{array,ulem}
\usepackage{cite}
\usepackage[caption=false,font=footnotesize,labelfont=sf,textfont=sf]{subfig}
\usepackage{textcomp}
\usepackage{stfloats}
\usepackage{url}
\usepackage{verbatim}
\usepackage{graphicx}
\usepackage{cite}
\usepackage{soul}
\usepackage{hyperref}
\usepackage{bm}
\usepackage{bbm}
\usepackage{subfig}
\usepackage{textcomp}
\usepackage{accents}
\usepackage{multirow}
\usepackage{bbding}
\usepackage{wasysym}
\usepackage{booktabs}
\usepackage{ulem}
\usepackage{geometry}
\newcommand{\hj}[1]{\textcolor{black}{#1}}
\newcommand{\xr}[1]{\textcolor{black}{#1}}

\newcommand{\TODO}[1]{\textcolor{black}{#1}}

\title{\LARGE \bf {PlanScope:} Learning to Plan Within Decision Scope\\
\textcolor{black}{for Urban Autonomous Driving}}

\author{
Ren Xin, Jie Cheng, Hongji Liu, and Jun Ma, \textit{Senior Member, IEEE}
	\thanks{This work was supported by the Guangzhou Education Bureau Project under Grant 2024312095.  \textit{(Corresponding author: Jun Ma.)}}
\thanks{Ren Xin, Hongji Liu, and Jun Ma are with the Division of Emerging Interdisciplinary Areas, The Hong Kong University of Science and Technology, Hong Kong SAR, China, and also with the Robotics and Autonomous Systems Thrust, The Hong Kong University of Science and Technology~(Guangzhou), Guangzhou 511453, China (e-mail: rxin@connect.ust.hk; hliucq@connect.ust.hk; jun.ma@ust.hk).}
\thanks{Jie Cheng is with the Department of Electronic and Computer Engineering, The Hong Kong University of Science and Technology, Hong Kong SAR, China (e-mail: jchengai@connect.ust.hk).}
}

\begin{document}
\maketitle

% 在端到端的自动驾驶训练过程中 我们提出一次进行长时序的规划是不合理的
% 有很多无法预见的未来状况会影响未来轨迹的生成
% 比如说突然出现的障碍物或者突然变化的交通信号灯
% 这些往往会导致细致且迅速的驾驶行为变化
% 同时未来轨迹会有一个大致的方向性
% 如沿着参考线行驶，或者避开静态障碍物等
% 本文提出了一些新的方法来解决这个问题

% 猜想
% 不同响应频率的速度规划结果应该与不同scale的场景信息相关
% L2 loss无法有效捕捉到高频变化之间的差异
% 此外参照diffusion model模型进行多步迭代或许可以提高结果的准确性
\begin{abstract}
\TODO{
In the context of urban autonomous driving, imitation learning-based methods have shown remarkable effectiveness, with a typical practice to minimize the discrepancy between expert driving logs and predictive decision sequences.} 
% Driving logs typically contain sudden obstacles or rapidly changing traffic signals, requiring short-term, responsive decisions. 
% Meanwhile, vehicle trajectories represent long-term decisions, such as maintaining a reference lane or avoiding stationary obstacles.
\TODO{
% Our hypothesis suggests
As expert driving logs natively contain future short-term decisions with respect to events, such as sudden obstacles or rapidly changing traffic signals. 
We believe that unpredictable future events and corresponding expert reactions can introduce reasoning disturbances, negatively affecting the convergence efficiency of planning models. 
At the same time, long-term decision information, such as maintaining a reference lane or avoiding stationary obstacles, is essential for guiding short-term decisions.
Our preliminary experiments on shortening the planning horizon show a rise-and-fall trend in driving performance, supporting these hypotheses.}
\TODO{
Based on these insights, we present PlanScope, a sequential-decision-learning framework with novel techniques for separating short-term and long-term decisions in decision logs. To identify and extract each decision component, the Wavelet Transform on trajectory profiles is proposed. After that, to enhance the detail-generating ability of Neural Networks, extra Detail Decoders are proposed. Finally, to enable in-scope decision supervision across detail levels, Multi-Scope Supervision strategies are adopted during training. 
The proposed methods, especially the time-dependent normalization, outperform baseline models in closed-loop evaluations on the nuPlan dataset, offering a plug-and-play solution to enhance existing planning models.
}

% This framework employs wavelet transformation based log preprocessing with an effective loss computation approach, rendering the planning model only sensitive to valuable decisions at the current state. 
% Since frequency domain characteristics are extracted in conjunction with time domain features by wavelets, decision information across various frequency bands within the corresponding time horizon can be suitably captured. 
% Furthermore, to achieve valuable decision learning, this framework leverages a transformer based decoder that incrementally generates the detailed profiles of future decisions over multiple steps. 
% Our experiments demonstrate that our proposed method outperforms baselines in terms of driving scores with closed-loop evaluations on the nuPlan dataset.
% 在 nuplan 任务中，我们应该是首先意识到这个问题并提出解决方法的工作。
% The source code and associated videos are available at \url{code_hidden} and \url{https://youtu.be/NxIIHp8bV1s}, respectively.
The source code and associated videos are available at \url{https://github.com/Rex-sys-hk/PlanScope}.

% \url{https://github.com/Rex-sys-hk/PlanScope}.
\end{abstract}

% TODO： 介绍我们方法中对动作sequence学习的普遍性
\section{Introduction}

\begin{figure}[!t]
  % \vspace{2em}
  \centering
  \includegraphics[width=\linewidth]{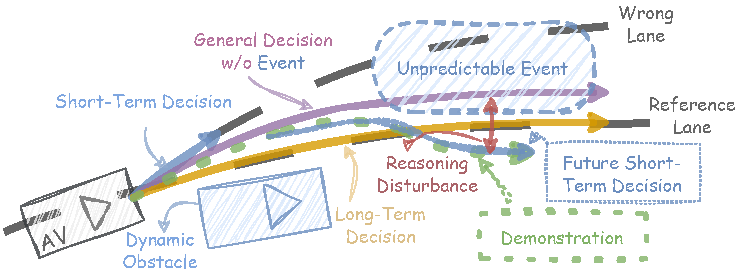}
  \vspace{-2em}
  \caption{
    In the scenario presented, the ego Autonomous Vehicle~(AV)~(black rectangle) is driving on a structured road with dynamic obstacles~(blue rectangle). Its long-term decision~(orange arrow) is characterized by adherence to the reference line~(black long dash) for driving. 
    Conversely, the short-term decision~(blue arrow) involves evading the approaching dynamic object through the application of lateral velocity. 
    Long-term and short-term decisions constitute the general decision~(purple arrow) at the moment. 
    However, the demonstrated trajectory~(green dashed arrow), which incorporates future short-term decisions~(blue dashed arrow) to avoid unpredictable events~(blue dashed area), is counterintuitive in the current context and introduces unexpected reasoning noise~(red double-headed arrow).
    % The decision scope mechanism deliberately focuses on a limited horizon of short-term decisions while maintaining a broader horizon for long-term decisions, thereby mitigating the effect of unpredictable decisions in the demonstrated trajectory.
    % excluding the unpredictable decision scope.
    }
    \label{fig:concept}
    \vspace{-2em}
\end{figure}

\TODO{Imitation learning has played a pivotal role in advancing the progress of urban autonomous driving~\cite{chauffeurnet, il_survey}.}
In particular, learning from driving logs of experts is broadly adopted to supervise neural networks for the planning task.
During the training process, distance error is commonly employed, which is to measure the discrepancy between the planned state sequence and the expert demonstrations, thereby optimizing parameters of the neural network~\cite{lbp_survey, tuplan, pluto, pip, trajectron++}. 
Yet, when training learning-based planning systems, it is not \TODO{absolutely} reasonable to set the logged expert states as the target sequence, because future events are inherently uncertain, and potential disruptions such as unexpected obstacles or fluctuating traffic signals can significantly alter the trajectory planning.
These unpredictable elements necessitate agile and nuanced adjustments to driving behavior. 
% 甚至是一些看似可以预测的事情，比如他车的交互行为或者未来轨迹，其实也是难以准确预测的。
% 例如在博弈模型中，只有当他车对我们的行为做出一定反应后，在才能准确得到他车的博弈模式。
% 否则我们只能在一个预设普适模型预测他车的交互形式
Some seemingly predictable decisions, such as behaviors in Game Theory models~\cite{game_zhou, gameformer} with multiple Nash Equilibria and switching traffic lights, are difficult to accurately predict.
% \sout{For example, when modeling the interactions by non-cooperative game theory, the decision planner converges only when observing other traffic participants respond.
% Otherwise, one can only avoid interactions with others to evade falling into disadvantageous states.}
% Otherwise, one can only predict the interaction \hl{form of other vehicles in a preset general interaction model parameter.}
Meanwhile, a short-sighted planner is also not desirable, 
as the planner also needs to keep its long-term decision in consideration when choosing sudden maneuvers.
% because the planner also needs to consider whether it can continue to drive smoothly on the road after the emergency obstacle avoidance behavior, and should not get stuck in obstacles or merge into lanes \hl{that it should not.}
% Therefore, future trajectories need to exhibit short-term decisions, such as following a reference lane or evading stationary obstacles.
Essentially, the state sequence of experts recorded in the log is a combination of short-term and long-term decisions under different scopes.
A diagram of this concept is shown in Fig.~\ref{fig:concept}.
% This understanding is vital for developing autonomous driving systems that can safely and effectively navigate complex and dynamic environments.
Due to the lack of observable information, long-term decisions are more susceptible to unpredictable events compared to short-term decisions. In learning targets like Average Distance Error~(ADE), waypoints near the given current state only account for a small proportion of error compared to remote ones. This will result in a situation where unpredictable events occurring at the distant future affect current short-term decisions unexpectedly. This issue is evident during the training phase of the planning model, where it can lead to uneven convergence priorities or errant convergence over time. 
% In this way, Neural Network~(NN) is not able to catch decision features in first several time steps, whose accuracy is of vital importantance in exacution, thereby leading to the deterioration of the planning performance.
\TODO{In summary, when training from expert demonstrations, trajectories inherently mix short-term reactive maneuvers and long-term directional decisions. Such entanglement often causes planning models to overfit to static future patterns and respond suboptimally to unexpected events.}
Therefore, considering the decision scope~(DS) in the model training phase is crucial for making rational and efficient planning decisions.
\TODO{This work addresses this problem by introducing learning-based decision-scope-aware trajectory planning for urban autonomous driving.}

Addressing the decoupling issue of information from various decision scopes for imitation learning presents significant challenges.
% within the learning-from-demonstration framework presents significant challenges. 
Traditional planning methods divide the problem into distinct layers: path planning, behavior planning, and motion planning. 
Specialized modules handle each of these layers. 
In contrast, imitation learning approaches often focus on learning future trajectories of a fixed length. 
This focus arises because nuanced adjustment components and long-term decision components are challenging to separate from log replay. 
One potential solution is to employ manual annotation for driving behavior reasoning. 
This method can help analyze both long- and short-term decisions, allowing the identification of behavior duration and start times.
However, this approach entails a substantial workload and poses scalability challenges. 
Consequently, finding a cost-effective way to decouple decision scopes in the age of learning-based planning remains a complex issue under exploration.

\TODO{
To extract decision components from expert trajectories in an unsupervised manner, we propose using Wavelet Transform~(WT) for temporal-frequency decomposition, where short-term decisions correspond to higher-frequency bands and long-term decisions to lower ones. After identifying these components, we develop Detail Decoding~(DD) structures to enable Neural Networks~(NNs) to generate detailed outputs. These NN-generated details are then supervised using Multi-Scope Supervision~(MSS) techniques, including leveraging weighted loss and decomposed expert driving logs for supervision.
}
Upon these designs, experimental validations of these strategies are conducted to ensure a comprehensive and practical exploration. 
Through our experiments, the most effective method is identified for enhancing model performance in closed-loop evaluations.
% \textcolor{blue}{To address aforementioned challenges, we propose three directions of solution, one is to weight the time dimension of loss, let the tracks be monitored by decisions of different scales, and let the model learn to refine the trajectories from coarse to fine.
% For time dpendent weight, we put forward three ways: Time Truncation, Uncertainty dependent weight and Time Dependent Normalization. These weights are added to the calculation of planning regression loss.
% For the decomposition of Expert Trajectory, we propose that the model can learn decisions of different scopes by using downsampling and truncation, or extract components of different frequencies from the expert trajectory by using wavelet decomposition and truncate them according to planning requirements.
% For the model structure of generating different scope decisions, we also tried two ways of one-time generation and iterative generation.
% We have carried out detailed experiments and analysis on the combination of the above methods, and effectively improved the performance of the model in the closed-loop evaluations.
% }
Our contributions are fourfold:
\begin{itemize}
  % \item We introduce PlanScope, a framework for formulating problems on training the sequential decision-making model, and demonstrate its impact by closed-loop evaluations.

  % \item We explore a variety of methods under PlanScope framework in the aspects of learning weight calculation, trajectory decomposition, and the modeling structure of detail generation. Through comparative experiments, the most effective strategies are identified for enhancing trajectory execution performance.

  % \item We extensively evaluate our method on the nuPlan~\cite{nuplan} dataset, and the results demonstrate that it achieves superior driving scores compared to baseline methods in closed-loop none-reactive simulations~(CLS-NR). 

  \item \TODO{We propose PlanScope, a decision-scope-aware planning framework that explicitly models the relationship between temporal horizon and planning performance.}
  \item \TODO{We propose and validate the functionality of adopting WT to extract and analyze decision components in different frequency bands from expert demonstrated trajectories.}
  \item \TODO{We develop DD structures that are able to recursively refine trajectory velocity details across multiple temporal levels.}
  \item \TODO{We design novel MSS strategies based on wavelet decomposition and weighted loss to supervise learning at different motion frequencies and time horizons.}
\end{itemize}
% \TODO{At last, we demonstrate, through closed-loop evaluations on nuPlan, that explicit scope modeling significantly improves planning safety and efficiency.}

% 并且可以通过希尔伯特变换得信号的波速或者相位信息
% And the signal's wave speed or phase information can be obtained through the Hilbert transform.
% % 此外我们猜测L2 loss可能无法有效捕捉到高频变化之间的差异，我们提出使用频域的损失函数来计算损失
% In addition, we speculate that L2 loss may not effectively capture the differences between high-frequency changes, and we propose to use a frequency domain loss function to calculate the loss.
% 最后我们参照diffusion model模型进行多步迭代来提高结果的准确性
% Finally, we refer to the diffusion model to perform multistep iterations to improve the accuracy of the results.
% 我们开发了一个新的解码框架来实现这一目标

\section{Related Works}
% 介绍本文可能有的普适性，与端到端的可结合性
% \subsection{Modern Decoder Architectures}

\subsection{Learning Strategies on Making Sequential Decisions}
\TODO{
Sequential decision-making can be encapsulated with the framework of Partially Observable Markov Decision Processes~(POMDPs)~\cite{pomdp}. 
In this framework, the ego agent must make decisions based on environmental observations, anticipate subsequent states following its actions, and maximize the cumulative reward. 
Temporal difference~(TD) together with Bellman Function formulates the basis of RL.
% ~\cite{RL_intro}. 
% such as Deep Q-Networks~(DQN)~\cite{dqn} and Proximal Policy Optimization~(PPO)~\cite{ppo}, are widely used to model value functions and optimizing policies.
In terms of RL for autonomous driving, it requires exponentially increasing search volume to find a feasible solution.
% Which consequence in exploring optimized policies in high-dimensional state spaces is computationally intensive and attributes high real-world or simulator trial cost.
This results in significant challenges when exploring optimized policies within high-dimensional state spaces, leading to substantial computational intensity. Consequently, this necessitates a high cost in terms of real-world or simulation trials.
Although modern learning strategies~\cite{SAC, A3C, TRPO} significantly improve exploring efficiency and optimality, they still require millions of trials and are still impractical for real-world implementation for autonomous driving because of the safety issue.}
% Given the world model (also known as the transition function) and the constraint of limited observations, the optimal policy can be derived by exploring the state space and resolving the Bellman equation.}

% \sout{Initially, it meticulously predicts the agent's reachable area at 0.1-second intervals for the first 0.5 seconds. 
% Subsequently, for the remaining 3.5 seconds, it utilizes an adaptive filter to depict potential vehicle positions. }
\TODO{
Imitation Learning~(IL) is an alternative approach that learns policies directly from expert demonstrations, thereby circumventing the need for explicit world models and reward functions~\cite{il_survey, chauffeurnet}.
However, existing IL methods often overlook the temporal and causal relationships inherent in sequential decision-making processes.
Whole-trajectory imitation typically treats all trajectory points equally, disregarding the varying significance of decisions made at different time scales within recognized driving patterns or history.
This oversight can lead to suboptimal performance, particularly in complex scenarios where decisions at different time scales have varying levels of importance.
Essentially, methodologies for solving these problems are under exploration.
Data augmentation, reason labeling, and scaling up the dataset are commonly used to alleviate the problems of distribution shift~\cite{pluto}.
Inductive bias can be introduced to alleviate these problems to some extent, like considering the reference lane as a strong prior~\cite{mpnp, tuplan}, or using an additional module to predict the traveling target first~\cite{TNT, HOME, GOHOME}.
But a generalizable method for extracting in-scope decisions from expert demonstrations with less intensive labor remains underexplored.
}

\subsection{Hierarchical Paradigm for Problem Simplification}

In the context of traditional rule-based autonomous driving systems, in-time reaction and predictive decision are both required, making it challenging to determine the optimal policy sequence based on POMDP in real-time~\cite{magic}.
Hence, under traditional planning frameworks, hierarchical structure is typically employed to streamline computations~\cite{epsilon, enhancing}. 
In the hierarchical structure, the planning problem is decomposed across various frequency bands, each with distinct hierarchical tendencies. 
For large-scale long-term decisions, one might opt for a lower resolution and focus on static objects by constructing maps~\cite{10588820, SDMAP}. 
Conversely, within the smaller-scale areas identified by the large-scale search, the planner can leverage higher accuracy and decision correcting frequency to deal with dynamic objects control inaccuracies~\cite{safe_ilqr, MPCC}.
% In~\cite{safe_ilqr}, the authors adapt this concept to the prediction phase and segment the forecast for the subsequent four seconds into two stages. 
% This approach allows the agent to maintain safety at a fine scale while preserving adaptability at a broader scale.

\TODO{The underlying mathematical principles can be traced to the Taylor expansion, which approximates complex polynomials with lower-order, linearized functions near a reference point~\cite{mpc}.
% By truncating the series at different orders, one can derive approximations that capture varying levels of detail.
% This concept is analogous to hierarchical planning, where decisions are made at different levels of granularity.
In our work, we draw inspiration from this framework to develop a learning-based scope-aware planning model that effectively facilitates long-term stability and short-term flexibility. 
}
% \TODO{Should introduce why we need signal decomposition}

% The local trajectory planning process can be modeled as mapping from the driving context to the future trajectory under the condition of routing result form higher level of coarse route network and our work targets further leverage this paradigm in learning-based planning in local planning modules towards better safety and efficiency.

\subsection{Signal Time-Frequency Analysis}

\TODO{Expert driving logs contain various decision components, and extracting these components to supervise planning model outputs is a promising strategy for enhancing performance. Signal analysis has long been used to extract meaningful information from complex data. Analytical tools like the Fourier Transform (FT), Laplace Transform (LT), and Hilbert Transform (HT) are commonly employed for frequency-domain analysis~\cite{FT, FFT, LT, HT}. 
For time-frequency signals, Short Time Fourier Transform (STFT) and WT are widely used to analyze non-stationary signals, with WT offering multi-resolution analysis that adapts to the signal characteristics, making it effective for capturing transient features~\cite{STFT, wavelet}.}
WT has been applied in autonomous driving, such as in~\cite{wave_categorizing} for categorizing car-following behaviors, preprocessing signal approximations, and identifying driving patterns through clustering. In~\cite{wvelet_id}, vehicle state records are analyzed using WT for driver identification, achieving nearly 100\% accuracy with classifiers like XGBoost. In~\cite{wavelet_map}, wavelet-based decomposition of high-resolution maps improves path and motion planning. 
\TODO{These applications demonstrate the effectiveness of WT in extracting features across diverse domains. 
Inspired by these works, we propose using WT to extract valuable information from expert trajectories for supervising decision-making across different time horizons.}

% wavelet 与 fourier 的区别
% wavelet 在模式识别中的应用
% wavelet 在去噪中的应用
% wavelet 的种类，母小波的种类

\section{Methodology} 

\begin{figure*}[ht]
    \centering
    \includegraphics[width=0.7\linewidth]{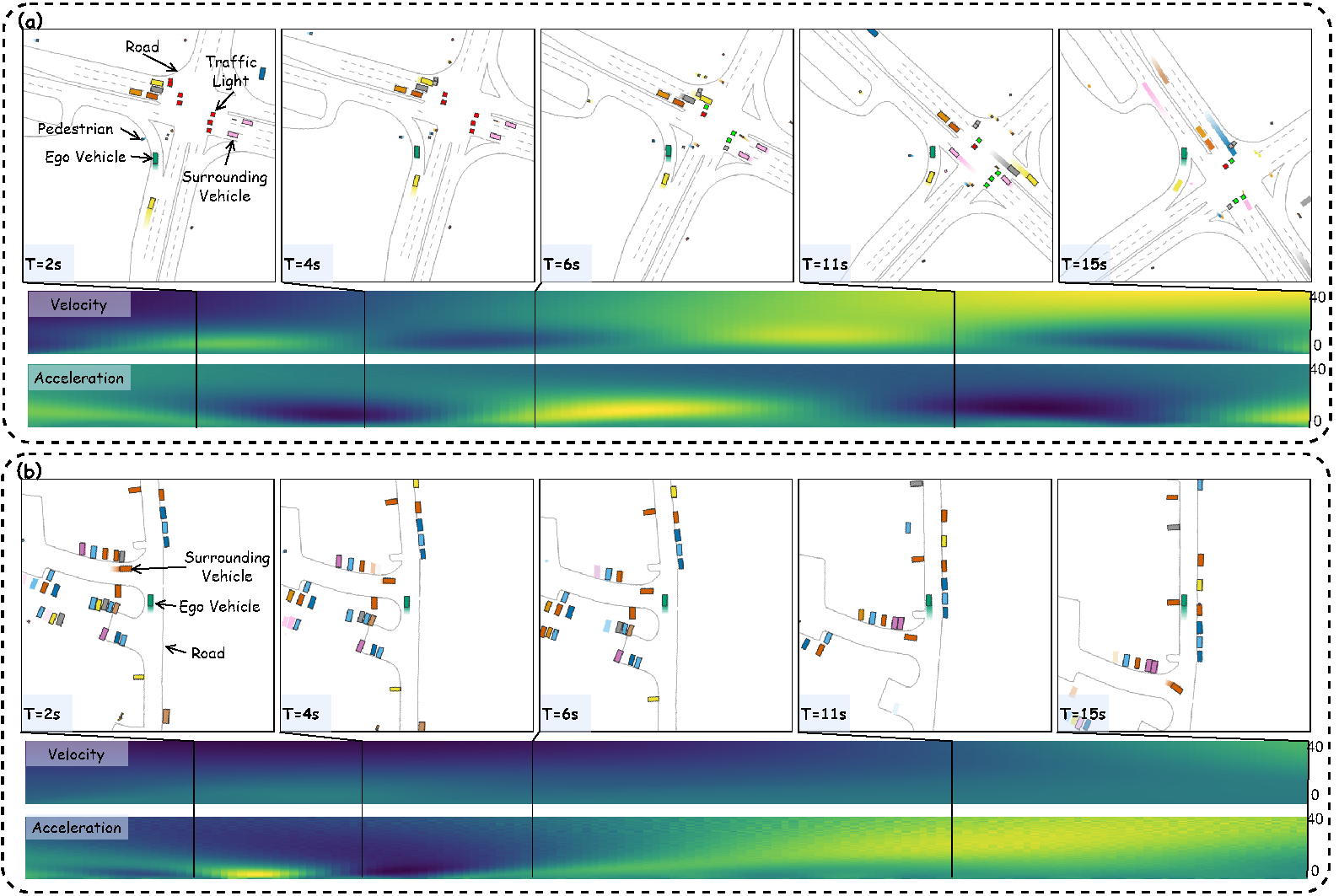}
    \vspace{-1em}
    \caption{Typical scenarios involving multi-scope decisions. 
    (a) The long-term decision is a left turn. Between T = 2\,s and T = 4\,s, the vehicle stops to yield to pedestrians, then accelerates at T = 6\,s to complete the turn. From T = 6\,s and T = 11\,s, it stops again upon observing the green light for cross traffic.
    (b) The ego vehicle first accelerates, then decelerates to secure the right of way. After observing another vehicle stop at T = 4\,s, it gradually resumes acceleration.
    CWT of expert velocity and acceleration profiles follows below.
    In scenario (a), the area where the scale of wavelengths larger than 20 indicates a general speed increase as the vehicle transitions from a branch road to a main street. At smaller scales, two deceleration segments correspond to two reactive yielding maeuvers.
    In scenario (b), acceleration decomposition reveals an accelerate-then-deccelerate decision around T = 4\,s, and a gradual speed-up around scale 25 between time steps 8-15. 
    % Noise in the expert trajectory appears at scales near 0.5.
    The colorbar for coefficient amplitude is omitted since it is unnecessary for qualitative analysis.
    }
    \label{fig:cases}
    \vspace{-1em}
\end{figure*}

\subsection{Scenario Analysis}

In Fig.~\ref{fig:cases}, we illustrate two typical cases involving multi-scope decisions to further elaborate on the proposed concept. Continuous Wavelet Transform~(CWT) using the Gaussian basis function is applied to capture driving maneuvers for the two cases.
\TODO{We choose CWT over Discrete Wavelet Transform~(DWT) to smoothly capture the changing patterns of coefficients across small to large scales. Because the Gaussian mother wavelet is a simple and widely adopted single-peak function, it can more effectively reflect the variance in trajectory profiles. }
From these two cases, we can clearly observe the additivity of self-driving maneuvers and unpredictability of certain traffic participants.

By generating trajectories in a multi-scope paradigm, the model at each level should be simplified and converge with fewer samples. 
If we decompose a trajectory into three levels with three times the length, the $n$ individual model cases should be simplified from $O(n\times3n\times9n)$ to $O(3n)$.

% \hl{weighgt is a variant of multi task, exonential because of self evolution}

% Our research is focusing on urban autonomous driving problem.
% Autonomous driving Vehicle~(AV) is learning to generate future trajectory~($\mathcal{T}$) under traffic-related contexts $C_t$ at a time step, including Dynamic Agents~(A), Static Objects~(O), and High-Definition Map~(M) information. 
% Generated items at time step t~($G_t$) also include motion prediction~($\mathcal{P}$) of other agents, multimodal trajectory score~($\mathcal{S}$) and trajectory details~($\mathcal{D}$) in different levels.
% The framework can be mathematically formulated as:
\subsection{Problem Formulation \& System Overview}

\begin{figure}[t]
\centering
\vspace{0.5em}
\includegraphics[width=\linewidth]{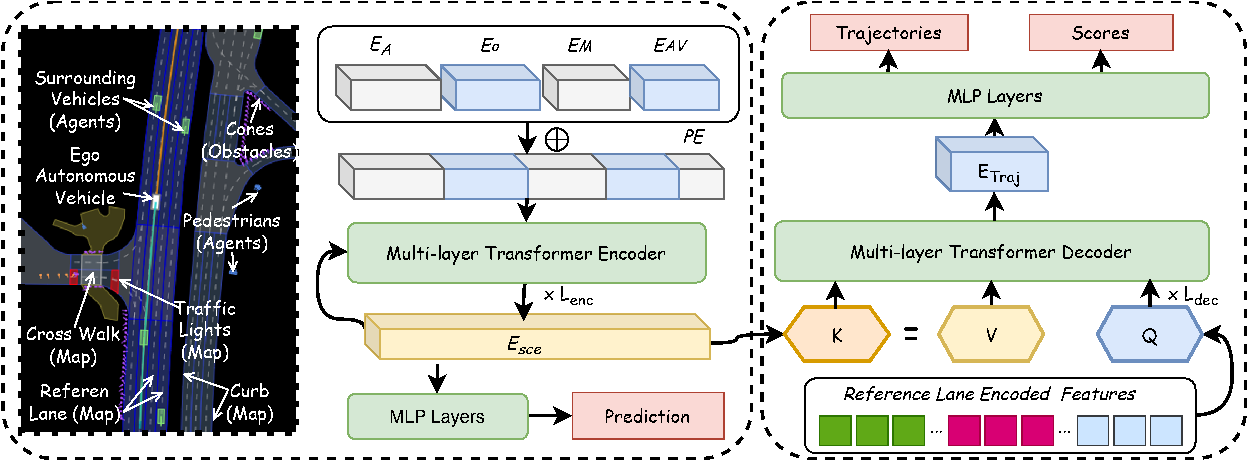}
\vspace{-2em}
\caption{
% The overall model framework. The driving context including Ego Vehicle, Dynamic Agents, Static Objects, High-Definition Map, Traffic Sign and Signals are first encoded respectively and then concatenated together with positional encoding~(PE). The encoded context is then fed into the Multi-layer Transformer encoder by $L_{enc}$ times to learn the spatial-temporal dependencies. The scenario encoding are first decoded as agent predictions, which is to supervise the model comprehensively understand the driving context like world model. Reference lane encoding and anchor free variables are concatenated as query to the scenario encoding. The decoder output generate residual latent variables. Latent variable of each decoder layer is logged as details of each layer and summed up with reference query as new query of next layer. The generated trajectories are then used as a reference by the LQR trajectory follower.
The overall model framework initiates with the individual embedding of driving context elements, including the \TODO{Dynamic Agents~(A), Static Objects~(O), Autonomous Vehicle~(AV), together with High-Definition Map, Traffic Signs, and Signals~(M)}, followed by their concatenation with positional embedding~(PE)~\cite{pe}. This concatenated context is subsequently processed through a Multi-layer Transformer encoder, iterated $L_{enc}$ times. The scenario embedding \TODO{serves as Key~(K) and Value~(V) of the transformer decoder and} is decoded to formulate agent predictions, which serve to supervise the model's understanding of the driving context comprehensively.
The reference lane embedding and anchor-free variables are combined to form a \TODO{Query~(Q)} for the scenario embedding. 
% The decoder's output generates latent variables, with each layer's latent variable logged as that layer's detail embedding.
}
\label{fig:scope_framework}
\vspace{-1.5em}
\end{figure}

Our research focuses on improving the trajectory quality generated by planning systems for urban autonomous driving, thereby contributing to driving efficiency and safety. 
At each time step, the autonomous vehicle is tasked with planning a future trajectory \( \mathcal{T} \) under various traffic-related contexts \( C_t \). 
\TODO{Each trajectory point has six channels:~\([p_x,p_y,\cos \eta,\sin \eta,v_x,v_y]\).
To ensure continuity in angle representation, the heading orientation is encoded as $(\sin \eta, \cos \eta)$, which avoids the periodic discontinuity at $\pm \pi $.}
Driving contexts encompass dynamic agents~(A), static objects~(O), and Vectorized High-Definition Map~(M) information. 
Generated outputs at time step \( t \), denoted as \( G_t \), include motion predictions \( \mathcal{P} \) of other agents, a multi-modal trajectory score \( \mathcal{S} \), and detailed trajectory information \( \mathcal{D} \) at different levels of detail.
A basic formulation of this framework can be represented as follows:
\begin{equation*}
    \begin{aligned}
        C_t &= \{A_t, A_{t-1}, \ldots, O_t, M_t, AV_t\}, \, G_t = \{\mathcal{T}, \mathcal{P}, \mathcal{S}, \mathcal{D}\}, \\
        G_t &= f(C_t|\theta),\, \text{w.r.t.} \, \theta = \arg\min_{\theta} \mathcal{L}(G_t, \hat{G}_t),
    \end{aligned}
\end{equation*}
where $f$ denotes the Neural Network~(NN) model, $\theta$ is the model parameter, $\mathcal{L}$ is the loss function, and $\hat{G}_t$ is the ground truth of $G_t$ from logged context.

% Regular model introduction
% The multimodel trajectory decoder is actually exploring more possibility of the future trajectory.
% 其他模型训练中常用的loss也会被用于我们模型的训练
% As shown in Fig.~\ref{fig:scope_framework}, the left encoder part is quite ordinary and mostly same as~\cite{pluto}, which is one of state-of-the-art prediction and planning model, while the right decoder part is carefully designed by us. 
% The generated planning result is then utilized as a reference by LQR trajectory follower.
% Rule-model hybrid post-processing module is eliminated in our experiments to reduce the bias of human defined rules.
% In th training process, other commonly used losses, including multimodal classification loss~($L_{cls}$), regression loss~($L_{reg}$), and collision loss~($L_{coll}$), are also used to supervise trajectory generation.
% Inspired by~\cite{detr, velf} and~\cite{mpnp}, the query of the decoder is designed as the concatenation of the reference lane encoding and anchor-free variables.
% The reference lane features are used as query $Q_0$ to the scenario encoding in multi-layer self-transformer operations, to generate residual quantities of $Q$ in next layer. 
% More information about how to get supervision signal and how to decode the detail information are introduced in Section~\ref{sec:decomposition} and Section~\ref{sec:iterative}.
% \TODO{Following the structure of driving models,} 
Driving contexts are encoded as \(E_A, E_o, E_{M}, E_{AV}\) by Feature Pyramid Network~(FPN)~\cite{plantf}, Multi Layer Perceptron~(MLP), PointNet~\cite{pointnet} based vector encoder and State Dropout Encoder~(SDE)~\cite{fmae}, respectively.
% As depicted in Fig.~\ref{fig:scope_framework}, the encoder section of our framework is conventional and similar to that of the baseline prediction and planning model described in~\cite{pluto}.
% We meticulously design the decoder module. 
% It generates trajectories that are subsequently employed as references by a Linear Quadratic Regulator~(LQR).
% To minimize the influence of human-defined rules, we have eliminated the rule-model hybrid post-processing module in our experiments. 
% During the training phase, various commonly utilized auxiliary losses are integrated to supervise the trajectory generation. 
As shown in Fig.~\ref{fig:scope_framework}, the encoder is structured as a multi-layer transformer, which is widely adopted in~\cite{tuplan, pluto}.  
% of our framework follows a conventional design, similar to the baseline prediction and planning model in~\cite{pluto}. 
The decoder is carefully designed as described in \ref{sec:dd}, generating trajectories that serve as references for a Linear Quadratic Regulator (LQR). 
% To reduce the impact of human-defined rules, we removed the rule-model hybrid post-processing module in our experiments. 
During training, common auxiliary losses are integrated to supervise trajectory generation.
The training losses include prediction loss \( \mathcal{L}_\text{pre}\), and collision loss \( \mathcal{L}_\text{col} \), which can be expressed by
\begin{equation*}
    % \begin{aligned}
    %     % \mathcal{L}_{cls} &= \\
    %     % \mathcal{L}_{reg} &= \\
        \mathcal{L}_\text{pre} = \operatorname{L1_{smooth}}\left(\mathcal{P}, \hat{\mathcal{P}}\right), \mathcal{L}_\text{col} = \frac{1}{T_{f}}\sum_{t=1}^{T_{f}}\sum_{i=1}^{N_{c}}\max\left(0, R_{c}+\epsilon-d_{i}^{t}\right), 
    % \end{aligned}
\end{equation*}
where \(\mathcal{L}_\text{pre}\) measures trajectory discrepancy between generated and that from the dataset, \(\mathcal{L}_\text{col}\) computes the invasion distance by \(N_c\) circle-based vehicle outlines over \(T_f\) future steps, \( R_c\) denotes the summed radius of ego and agent circles, \(\epsilon\) is the tolerance, and \(d_i^t\) represents the center distance at time \(t\).
Inspired by~\cite{detr, velf, mpnp}, the decoder query combines reference lane encodings with anchor-free learnable variables, forming the initial query \( Q_0 \) for multi-layer transformer decoding and subsequent residual updates of \( Q \).
During training, the final query $Q_{N}$ is then forwarded to the trajectory MLP decoder and scoring MLP decoder, yielding the multi-modal trajectory and the score by
\begin{equation*}
    \begin{aligned}
        \mathcal{T} &= \operatorname{MLP}(Q_{N}) ,\, \pi = \operatorname{MLP}(Q_{N}), \\
        \mathcal{L}_\text{cls} & = \operatorname{CrossEntropy}\left(\pi, \hat{\pi}\right),
    \end{aligned}
\end{equation*}
where \(\hat{\pi}\) is a one-hot vector indicating the trajectory closest to the logged trajectory.
% MSL does not directly target the detail information but functions as an auxiliary loss. 
% This approach is taken because, while wavelet decomposition is theoretically inversible, practical challenges such as data noise and boundary effects can impede complete reversibility. 
% Consequently, we employ MSL to facilitate the model's learning of detail information and to minimize the propagation of errors.
Finally, the overall training loss is
\begin{equation*}
    \mathcal{L} = \mathcal{L}_\text{reg} + \mathcal{L}_\text{cls} + \mathcal{L}_\text{pre} + \mathcal{L}_\text{col} + \mathcal{L}_\text{ds},
\end{equation*}
\xr{where $\mathcal{L}_\text{reg}$ is the weighted regression loss and $\mathcal{L}_\text{ds}$ is the loss to supervise learning within decision scope.}
More details on weights design, acquisition of supervision signals, and the decoding of detailed information are elaborated in Section~\ref{sec:weights}, Section~\ref{sec:decomposition}, and Section~\ref{sec:iterative}, respectively.

\subsection{\xr{Weighted Loss}}\label{sec:weights}

\begin{figure}[!t]
\centering
\includegraphics[width=0.9\linewidth]{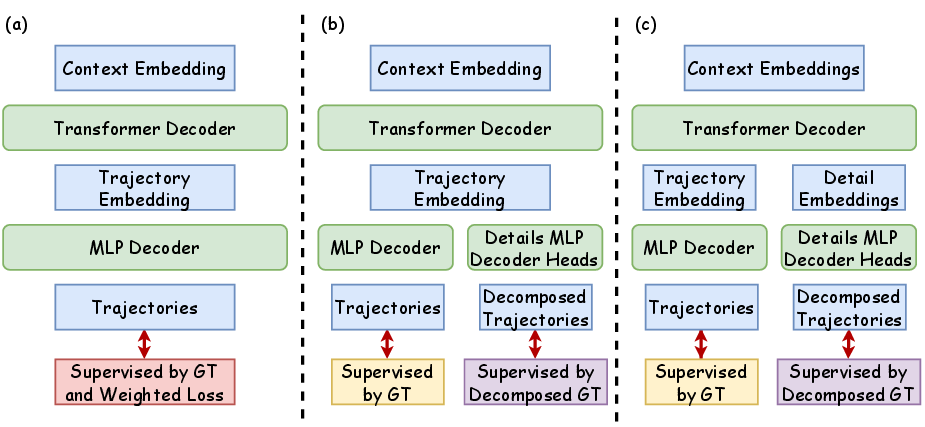}
\vspace{-1em}
\caption{
Multi-scope trajectory supervision is achieved in three forms: (a) time-weighted loss on the baseline; (b) decoding final embedding as both full and decomposed trajectories; (c) iteratively decoding multi-scope detail embeddings as corresponding trajectory details.
}\label{fig:scope_model_structures}
\vspace{-1.5em}

\end{figure}

To emphasize certain parts in demonstrated trajectories, it is an alternative to consider the precision of waypoints at each time step as separate tasks. 
Using time-dependent weight along the decision sequence is a promising way to balance convergence pace across tasks, as shown in Fig.~\ref{fig:scope_model_structures}(a).
The weighted regression loss can be calculated by
\begin{equation*}
    \begin{array}{c}
        \mathcal{L}_\text{reg}{}_t = \operatorname{L1_{smooth}}\left( \mathcal{T}_t, \hat{\mathcal{T}}_t \right), \\
        \mathcal{L}_\text{reg} = \frac{1}{T}\sum^{T}_{t=0} \mathcal{L}_\text{reg}{}_t \times w_t ,
    \end{array}
\end{equation*}
where $w_t$ can be designated by the following methods:
\subsubsection{\xr{Time Truncation}}
Only the loss of the first T time steps is considered. In implementation, the executed trajectory is shortened to remove the influence of uncontrolled waypoints
\begin{equation*}
    w_t = 
\left\{
    \begin{array}{ll}
                1, & t< T, \\
                0 ,& t\geq T.
    \end{array}
\right.
\end{equation*}  

\subsubsection{\xr{Decreasing Certainty}}
\xr{Those unpredictable disturbances might affect convergence capability on more important features and information for immediate decision. 
To describe this problem, we assume the sequential decision process conforms to the Gaussian Process~(GP).
% Predictable waypoints can be regarded as observed data with uncertainty in GP.
}
% GP is a nonparametric Bayesian method for modeling functions. 
% It assumes that any finite collection of function values follows a joint Gaussian distribution. 
% For trajectory planning of an autonomous vehicle, 
The trajectory points $\mathbf{y}(t)$ can be modeled as outputs of a GP, with the mean and covariance determined by a kernel function $ k(t_i, t_j) $, \hj{which} can be \hj{a} Radial Basis Function~(RBF).
Assuming the current state is an observable point, the variance of the predicted point at $t$ can be easily derived by
\begin{equation*}
% \[
\sigma_t^2 = \sigma_f^2 \left[ 1 - \exp\left(-\frac{(t - t_0)^2}{l^2}\right) \right],
% \]
\end{equation*}
where $\sigma_f^2$ is the function uncertainty, which can also be considered as the maximum uncertainty, $l$ is the time length scaler, $t_0$ is the observable current time, which is set to $0$ in the calculation. 
We define the uncertainty compensation as the deviation of maximum uncertainty:
\begin{equation*}
     \text{Compensation}(t) = \sigma_f^2 - \sigma_t^2.
\end{equation*}
Then, the calculation of weights to regularize uncertainty at each time point can be simplified as
% To normalize the significance of short and long-term decision significance, we \hj{employ} the reciprocal of uncertainty propagation as the weight for the distance error at \hj{the specific moment $t$ following}
\begin{equation*}
    w_t = \frac{1}{Z}\operatorname{exp}(-(\frac{t-t_0}{l})^p),
\end{equation*}
where $Z$ is the average weight. $p$ is the order of the time difference measurement, which is $2$ for RBF.
% , and $l$ is set to $e^\frac{1}{2}$ in our experiments.
% Due to the existence of dynamic constraints and reference lines, we believe that RBF might overestimate uncertainty. 
% Besides, we also attempted to adjust the order of the time difference measurement in RBF as
% \begin{equation*}
%     w_t = \frac{1}{Z}\operatorname{exp}(-\frac{||t-t_0||_p}{l^2}).
% \end{equation*}
% \xr{where $\alpha$ is the uncertainty growing speed coefficient, \hj{depending} on the designation of kernel function.}

\subsubsection{\xr{Time Dependent Normalization}}
% \xrnew{TODO, borrow idea from batch normalization, provement, to make loss evenly treated.}
% Normalization is a straightforward method 
\TODO{To convert the value of data to the same order of magnitude or limit it to a certain range, we borrow the idea of Batch Normalization~(BN). By time-dependent normalization,} the average error $e_t$ of each time step across the mini-batch can be dynamically calculated by
\begin{equation*}
    BN(e_t) = \frac{e_t-\mu_{B;t}^{e_t}}{\sigma_{B;t}} = \frac{e_t}{\mu_{B;t}^{|e_t|}}, \text{ when } \mu_{B;t}^{e_t}=0.
\end{equation*}
So, the distance error at each timestep $e_t$ can be normalized by multiplying by weights
\begin{equation*}
\begin{array}{c}
    w_t = \left(\frac{1}{B}\sum^{B}_{b=1}\mathcal{L}_{reg}{}_{b;t}\right)^{-1}, 
\end{array}
\end{equation*}
where $b$ is the index of the data item in the mini-batch and $B$ is the batch size.

\subsection{Expert Driving Logs Decomposition} \label{sec:decomposition}
In order to obtain decision components in the trajectory logs and selectively study them, we decompose them through mathematical analysis tools.
\TODO{By treating trajectory profiles as temporal signals, we use the Haar wavelet as the basis function for DWT. 
DWT is preferred over CWT due to its higher computational efficiency, facilitated by discrete downsampling at each wavelet scale, which is crucial for batch processing. 
The Haar wavelet is selected for its compact support, strong response to discontinuities, and identity property, making it well-suited to capture abrupt driving maneuvers like braking or lane changes, while minimizing bias.}
% Although smoother wavelets (e.g., Daubechies) capture gradual trends, they are less efficient for real-time applications. The horizon lengths (10-40 steps) used in our experiments correspond to 1-4 seconds at 10 Hz, matching typical short-term and mid-term decision windows in urban driving.}
% \sout{
% For example, the Haar wavelet, the simplest basis function of the wavelet transformation, enjoys widespread application in signal processing.}
% % Despite its simplicity, the Haar wavelet possesses several beneficial properties, including linear independence, compact support, and orthogonality.
% \sout{The basic functions of the Haar wavelet are}
% \begin{equation*}
%     \begin{aligned}
%         \phi(t) = \begin{cases} 1, & 0 \leq t < 1 \\ 0, & \text{otherwise} \end{cases},\, \psi(t) = \begin{cases} 1, & 0 \leq t < \frac{1}{2} \\ -1, & \frac{1}{2} \leq t < 1 \\ 0, & \text{otherwise} \end{cases},
%     \end{aligned}
% \end{equation*}
% \sout{where $t$ is the discrete time index.
% % 由公式可以看到，当t不属于0-1时，基函数值为0，即代表函数是紧支撑的
% We can see that when $t\not\in[0,1]$, the basis function value is 0, which means that the function is compactly supported.
% Among the basis functions, the scaling function \( \phi(t) \), often referred to as the father wavelet, serves as a low-pass filter within the wavelet transform process; the mother wavelet function \( \psi(t) \) is primarily utilized for extracting the high-frequency components of a signal. 
% Through the convolution of the original signal with these basis functions, the signal's high-frequency and low-frequency components are isolated. 
% }
In the trajectory decomposition context, the trajectory approximation coefficient \( \mathcal{T}_A \) and the detail coefficient \( \mathcal{T}_D \) can be derived by performing the following transformation over the time horizon
\begin{equation*}
    \begin{aligned}
        \mathcal{T}_{At} = \sum_{k} \mathcal{T}[k] \phi\left(2t-k\right),\, \mathcal{T}_{Dt} = \sum_{k} \mathcal{T}[k] \psi\left(2t-k\right),
    \end{aligned}
\end{equation*}
where \( \mathcal{T}[k] \) is the discrete trajectory profile, \( \mathcal{T}_{At} \) is the low-frequency approximation coefficient at time index $t$, \( \mathcal{T}_{Dt} \) is the high-frequency detail coefficient at time index $t$, \(k\) can be considered as the convolution handler.
\xr{For Haar wavelet decomposition, \( \mathcal{T}_{At} \) \hj{represents the} sum of \(\mathcal{T}[k]\) and \(\mathcal{T}[k+1]\) while \( \mathcal{T}_{Dt} \) is the product similarities bwtween \([\mathcal{T}[k],\mathcal{T}[k+1]]\) and \([1,-1]\).}
% In the one-dimensional Discrete Wavelet Transform (DWT), a signal \( x[n] \) is processed using two filters: the approximation filter \( h[n] \) and the detail filter \( g[n] \). These filters are applied to the signal to extract different frequency components. 
% The mathematical representation of this process is given by:
% \begin{equation*}
%     \begin{aligned}
%         a[n] &= \sum_{k=-\infty}^{\infty} x[k] h[n-k], \\
%         d[n] &= \sum_{k=-\infty}^{\infty} x[k] g[n-k],
%     \end{aligned}
% \end{equation*}
% where \( a[n] \) represents the approximation coefficients, capturing the lower frequency components or the smooth version of the signal, and \( d[n] \) represents the detail coefficients, capturing the higher frequency components or the oscillatory parts of the signal. 
% The index \( k \) is used for the convolution calculation, effectively shifting the mother wavelet by \( k \) when it is multiplied by the signal at each time step \( n \).
\begin{figure}[t]
    \centering
    \includegraphics[width=\linewidth]{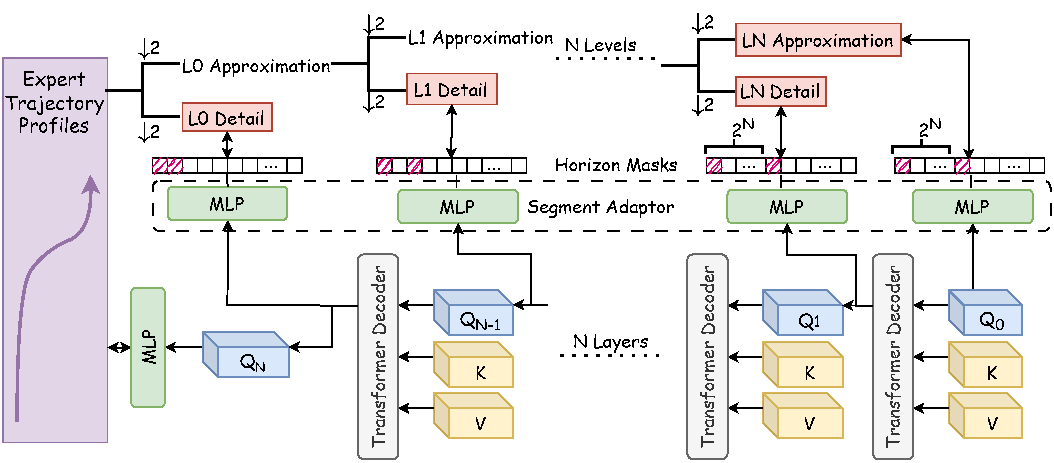}
    \vspace{-2em}
    \caption{
    % The structure of the iterative detail decoder. The reference lane encoding is used as the query for multi-layer self-attention operations. The detail information obtained at each iteration will be summed to the previous query as the next query to obtain next detail information. The first query and generated details are remapped by MLP and then compared with detailed of each detail level of ground truth trajectory. The initial query is used as an approximation. The subsequent velocity residual values will be supervised by the velocity detail at different levels. Specifically, the max counted horizon is the same for each level, the sampling interval of each detail level is twice as the previous level.
    The iterative detail decoder uses reference lane encoding as the initial query for multi-layer attention. At each iteration, extracted detail coefficients are combined with the previous query to refine further detail extraction. The initial query provides an approximation, and both it and subsequent generated details are processed via an MLP. The resulting detail embeddings are then mapped and compared against corresponding components at each level, with horizon masks configurable per level.
    }
    \label{fig:iterative_velocity_decoder}
    \vspace{-1.5em}
\end{figure}
% In practical applications, the filters \( h[n] \) and \( g[n] \) are typically finite in length and symmetric, which simplifies their implementation in digital signal processing systems. 
% These filters can be derived from a mother wavelet function \( \psi(t) \) and a scaling function \( \phi(t) \). 
% For a given wavelet \( \psi(t) \), the filter coefficients are calculated using the relationships:
% \begin{equation*}
%     \begin{aligned}
%         h[n] &= \sqrt{2} \psi(2t - n), \\
%         g[n] &= \sqrt{2} {(-1)}^n \psi(2t - n - \frac{1}{2}),
%     \end{aligned}
% \end{equation*}
% where \( t \) is the scale parameter, and \( n \) is the time index.
This decomposition process is called DWT, which can be recursively applied to the approximation coefficient \( \mathcal{T}_A \) to achieve multi-level $l$ decomposition. 
% Each level of decomposition results in a new set of approximation and detail coefficients, which represent the signal at progressively lower frequencies and higher frequencies, respectively. 
% This hierarchical decomposition is particularly useful for analyzing signals at different scales and for applications such as data compression, noise reduction, and feature extraction in various domains including image processing and time series analysis.
% 通过小波变换，我们可以得到不同level的子轨迹
% 这是一个类似于卷积的操作，在该尺度上更明显的特征被提取在该scale的频谱中
% 该成分被称为细节成分
% 通过多次小波变换，我们可以得到不同level的细节成分
% 而最后剩下的为相似性成分
% 在我们的实验中，历史轨迹也会被用于轨迹的分解，以尽量减少信号处理时边缘补值的影响
% 通过变换的轨迹每被提取一次细节，数据长度就会减少为原来的一半
% 我们选择haar小波作为小波基函数
By 
\begin{equation*}
    \left(\mathcal{T}_A^l, \mathcal{T}_D^l\right) = \operatorname{DWT}\left(\mathcal{T}_A^{l-1}\right),
\end{equation*}
we decompose trajectory profiles into $N$ levels. 
The profile in our experiments is selected to be the coordinate values of waypoints. %, each representing features of the trajectory at different scales. 
The approximation obtained in the last level is nominated as the preliminary decision.
With the extraction of a detail component in each level, the length of the data is reduced by half.

\subsection{Detail Decoding}\label{sec:dd}
Since the decomposed trajectory profile has been obtained, the question arises as to how it can be effectively utilized to guide the supervision of the model training process.
% \sout{One potential method is to reconstruct the whole trajectory by truncated detail profiles.
% Nevertheless, this method requires assumption of data out of horizon to be zero, which introduces more deterministic prior in supervision signal.}

\subsubsection{\xr{Multi-head Detail Decoder~(MDD)}}
 % This method leverages the function-fitting ability of MLP to keep gradients. 
As shown in Fig.~\ref{fig:scope_framework} and \ref{fig:scope_model_structures}\,(b), the transformer decoder layer generates trajectory embedding. 
\TODO{Benefits from the function-fitting ability of MLPs, each decoding head with an MLP is able to decode trajectory embedding to details of designated levels or the overall approximation without interference. This process can be expressed as}
\begin{equation*}
    \tilde{\mathcal{T}}_A^N = \operatorname{MLP}\left(Q_N\right), \, \tilde{\mathcal{T}}_D^l = \operatorname{MLP}_l\left(Q_{N}\right),
\end{equation*}
\TODO{where the tilde means the trajectory detail and approximation is predicted by the model.
This design aims to train MLPs as detail extractors across different levels, without introducing artificial intervention or associated bias.}

\subsubsection{Iterative Detail Decoder~(IDD)} \label{sec:iterative}

% Iterative velocity decoder structure

% 做了个小实验，multihead的效果好像也不差于IDD，难受
% Another potential method is to use multi-head framework, outputting a level of detail by each head.
% However, this framework is not likely to supervise the head of outputting of the final trajectory efficiently.
% 为了让模型能够更好地学习速度信息，我们设计了一个迭代的速度解码器，该解码器可以通过多次迭代来逐渐提取速度信息。
% 在解码过程中，我们将bottle neck 隐变量作为query进行多次自注意力运算
% 每次迭代得到的速度信息会与上次的query进行加和，再加入到下一次的query中，再得到更为细节的速度信息。
% To better learn the velocity information, we design an iterative velocity decoder that can gradually extract velocity information through multiple iterations.
% During the decoding process, we use the bottleneck hidden variable as the query for multiple self-attention operations.
% The velocity information obtained at each iteration will be added to the previous query and then added to the next query to obtain more detailed velocity information.

To facilitate the learning of detailed information, IDD is designed, which is capable of progressively generating details across multiple iterations as shown in Fig.~\ref{fig:scope_model_structures}\,(c). 
The architecture of the IDD is depicted in Fig.~\ref{fig:iterative_velocity_decoder}.
In the decoding process, the embedding of the reference lane serves as the query \(Q_0\), and scenarios encoding \(E_{sce}\) serves as the key and value for a series of self-attention operations. 
\xr{
% Different from widely adopted multilayer transformer decoder in NLP tasks, we hypothesize planning is a problem in continuous space, where accuracy is more important than logits. 
% So, information entropy losing hierachical tranformer is substituted. 
% The original embedding fully participate in the calculation in next layer. 
%Transformed detail token at that level are combined as new query seed. 
We hypothesize that the multi-layer transformer decoder serves as a trajectory refining process. This hypothesis is based on the observation that the layer normalization operation within each decoding layer reduces information capacity while enhancing accuracy. 
% because the layer normalization operarion in each decoding layer can be considered as a information capacity decreasing but accuracy increasing process.
}
The embeddings extracted at each iteration are accumulated with the previous query, which is then utilized as the subsequent query.
% \sout{This iterative accumulation allows for the acquisition of increasingly refined trajectory profile.} 
The process can be described as
\begin{equation*}
    \begin{aligned}
        % {Q}_{i-1}' &= \text{SelfAttn}\left(Q_{i-1}, \text{dim}=0\right), \\
        % \tilde{Q}_{i-1} &= \text{SelfAttn}\left(Q_{i-1}', \text{dim}=1\right), \\
        Q_{i} &= \text{TransformerDecoder}\left(Q_{i-1}, E_{\text{sce}}, E_{\text{sce}}\right).
    \end{aligned}
\end{equation*}
% 每次迭代得到的速度信息都会被记录下来
% 其中最初的query作为approximation
% 后续的速度残差值会作为不同level下的速度detail
% 这些速度信息经过全连接层处理后会与上面处理过的不同level的专家轨迹进行比较
% 不同次的残差值与一开始的query刚好与不同level的专家轨迹相对应
% 最终的query会被送入到全连接层中，并通过MLP decoding得到最终的规划结果
% 我们的detail loss并不会直接作用于速度信息，而是作为一种peripheral loss，用于辅助模型学习速度信息
% 这是因为虽然理论上小波分解的过程是可逆的，但是在实际应用中，由于数据的噪声和边界效应，小波分解的过程并不是完全可逆的。
% 因此，我们通过detail loss来辅助模型学习速度信息，来减少引入误差。
% The velocity information obtained at each iteration will be recorded.
% The initial query is used as an approximation.
% The subsequent velocity residual values will be used as the velocity detail at different levels.
% These velocity information will be compared with the expert trajectories at different levels after being processed by the fully connected layer.
% The residual values at different times and the initial query correspond exactly to the expert trajectories at different levels.
% The final query will be sent to the fully connected layer and the final planning result will be obtained through MLP decoding.
% Our detail loss does not directly act on the velocity information, but serves as a peripheral loss to assist the model in learning the velocity information.
% This is because although the process of wavelet decomposition is theoretically reversible, in practical applications, due to data noise and boundary effects, the process of wavelet decomposition is not completely reversible.
% Therefore, we use the detail loss to assist the model in learning the velocity information to reduce the introduction of errors.
% The iterative velocity decoder structure is shown in Fig.~\ref{fig:iterative_velocity_decoder}.
\xr{In the decoding process, \hj{the} transformed feature $ Q_i $ is generated during each iteration and recorded.
Logged $ \{Q_i\} $ is adapted by level-wise MLP decoders, i.e.,}
\begin{equation*}
\begin{aligned}
    % \tilde{\mathcal{T}}_A^N = \operatorname{MLP}\left(Q_0\right), \, \tilde{\mathcal{T}}_D^l = \operatorname{MLP}\left(Q_{l}\right), or \\
    \tilde{\mathcal{T}}_A^N = \operatorname{MLP}_A\left(Q_0\right), \, \tilde{\mathcal{T}}_D^l = \operatorname{MLP}_l\left(Q_{l}\right).
\end{aligned}
\end{equation*}
% The initial query $Q_0$ serves as the overall approximation of the trajectory profile. 
% Residual variables $\Delta Q_i$ serve as driving details at various levels. 
The length of \(\tilde{\mathcal{T}}\) is the same as the planning steps. To make sure the decoded details are of the same length as the driving details, downsampling is performed on $\tilde{\mathcal{T}}$ by 
\begin{equation*}
\begin{aligned}
    {\mathcal{T}}_A^N = \tilde{\mathcal{T}}_A^N \downarrow 2^N , \, {\mathcal{T}}_D^l = \tilde{\mathcal{T}}_D^l \downarrow 2^{l+1} .
\end{aligned}
\end{equation*}
These details are compared with the corresponding components of expert trajectories at multiple levels.
To get decision details in a proper scope, only a limited time horizon $H_l$ of $\mathcal{T}_D^l$ at level $l$ is within the decision scope, supervising model training by
\begin{equation*}  
\begin{aligned}
    \mathcal{L}_\text{ds} & = \\
    \frac{1}{N+1} & \left(\sum^{N}_{l=0} \left(\lVert \mathcal{T}_D^l [:H_l] - \hat{\mathcal{T}}_D^l [:H_l] \rVert_2 \right) + \lVert \mathcal{T}_A^N-\hat{\mathcal{T}}_A^N \rVert_2\right).
\end{aligned}
\end{equation*}

\section{Experiments}

\begin{table}[!t]
    % \begin{tabular}{cccc}
    %     \hline
    %     Direct Velocity Loss & Recursive Decoder & Horizon & CLS-NR Score \\ \hline
    %     &                   &         & 81.47\%    \\
    %     \Checkmark           &                   &         & 83.60\%    \\
    %     & \Checkmark        & 10      & 84.30\%    \\
    %     & \Checkmark        & 20      & 86.80\%    \\
    %     & \Checkmark        & 30      & 83.56\%    \\
    %     & \Checkmark        & 40      & 84.37\%    \\
    %     & \Checkmark        & all     & 82.90\%    \\ \hline
    % \end{tabular}
    \centering
    \vspace{0.5em}
    \caption{\raggedright {Preliminary Ablation Studies on Decoding and Supervision Strategies}}
    \label{tab:ablation}
    \vspace{-1em}
    % \setlength{\tabcolsep}{10pt}
    % \renewcommand{\arraystretch}{1.2}
    % \small
    % \begin{tabular}{@{}lccc@{}}
    %     \toprule
    %     Remark               & IDD       & DS Horizon & Score \\ \midrule
    %     No Additional Loss   & -         &    -     & 81.47\%      \\
    %     Decision Scope Absent& \Checkmark & 0       & 84.69\%      \\
    %     Position             & \Checkmark & 20      & 85.80\%      \\
    %     Direct Velocity Loss &  -         &  -      & 83.60\%      \\
    %     Velocity             & \Checkmark & 20      & 86.77\%      \\ \midrule
    %     Velocity             & \Checkmark & 10      & 85.85\%      \\
    %     Velocity             & \Checkmark & 30      & 85.76\%      \\
    %     Velocity             & \Checkmark & 40      & 84.37\%      \\ \bottomrule
    %     % Velocity             & \Checkmark & All     & 82.90\%      \\ \bottomrule
    % \end{tabular}
\begin{tabular}{@{}cc|cc|cc@{}}
\toprule
\multicolumn{2}{c|}{Detail Gen.}                      & \multicolumn{2}{c|}{Supervision}                                &                   & \multicolumn{1}{l}{} \\ \midrule
Iterative                & Detail Heads              & Horizon & DWT                       & Remarks           & CLS-NR               \\ \midrule
-                         & -                         & 80      & -                         & PLUTO, m6         & 83.28\%              \\ \midrule
-                         & -                         & 10      & -                         & Truncation        & 80.12\%              \\
-                         & -                         & 20      & -                         & Truncation        & 87.92\%              \\
-                         & -                         & 40      & -                         & Truncation        & 85.22\%              \\ \midrule
-                         & -                         & -       & -                         & Time decay        & 88.36\%              \\
-                         & -                         & -       & -                         & Time norm         & 87.97\%              \\
% -                         & -                         & -       & -                         & -                         & Time norm+contras & 92.22\%              \\ 
\midrule
% -                         & \Checkmark & 10      & -                         & \Checkmark & -                 & 85.52\%              \\
% \Checkmark & -                         & 10      & -                         & \Checkmark & -                 & 82.43\%              \\
-                         & \Checkmark & 10      & \Checkmark                         & Position          & 82.26\%              \\
\Checkmark & -                         & 10      & \Checkmark                         & Position          & 84.71\%              \\
-                         & \Checkmark & 20      & \Checkmark                         & Position          & 82.48\%              \\
\Checkmark & -                         & 20      & \Checkmark                         & Position          & 82.96\%              \\ \bottomrule
\end{tabular}
    \vspace{-2em}
\end{table}

\begin{figure*}[t]
    \centering
    % \vspace{0.5em}
    \includegraphics[width=0.95\linewidth]{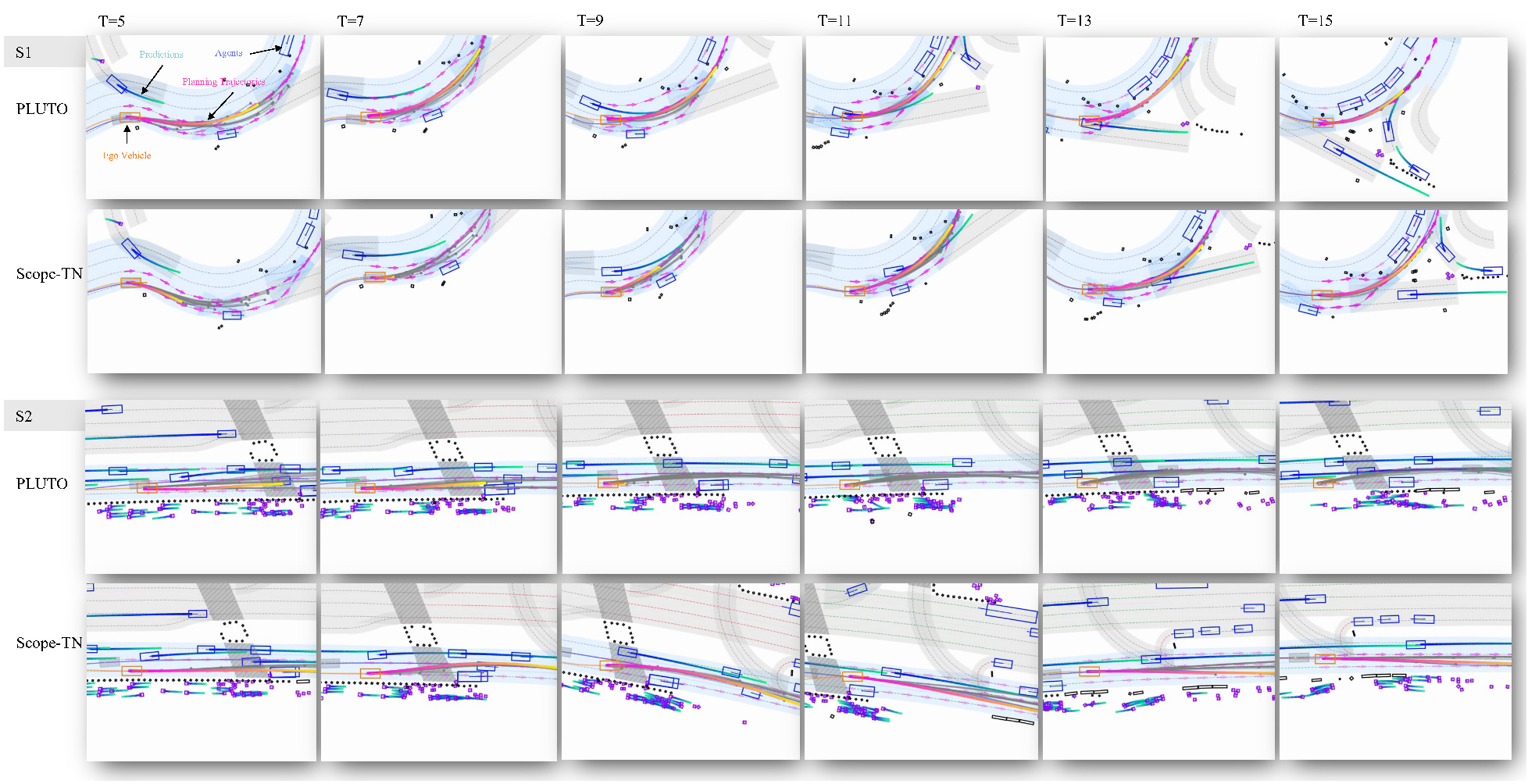}
    \vspace{-1.5em}
    % \vspace{-1em}
    \caption{The qualitative comparison is conducted between the standard PLUTO and PlanScope~(Scope) with time-dependent normalization~(TN), whose quantitative results are listed in Table~\ref{tab:sota}, fourth and third row from the bottom, respectively. In scenario \textbf{S1}, \textbf{PLUTO} fails to make a timely response to the impending collision in the short term, but chooses to continue along the long-term decision. In contrast, \textbf{Scope} makes a timely yielding maneuver. In scenario \textbf{S2}, \textbf{PLUTO} adopts a conservative decision due to the high uncertainty of the future trajectory caused by interaction with the rear vehicle in a long time horizon. In contrast, \textbf{Scope} adopts a human-like lane change decision.}
    \label{fig:qualitative}
    \vspace{-1em}
\end{figure*}

\begin{table*}[!ht]
    % \centering
    \centering
    % \vspace{0.5em}
    \caption{\raggedright Comparison with Baseline Methods}
    \label{tab:sota}
    \vspace{-1em}
% \usepackage{booktabs}
\begin{tabular}{@{}lcccccccll@{}}
\toprule
Planner            & Comfortness                    & Progress                       & w/o Collision                  & in Speed Limit                 & Drivable                       & TTC                                                 & CLS-NR                         & \multicolumn{1}{c}{CLS-R}                                               \\ \midrule
 \color[HTML]{808080}Log-Replay         & {\color[HTML]{808080} 99.27\%} & {\color[HTML]{808080} 98.99\%} & {\color[HTML]{808080} 98.76\%} & {\color[HTML]{808080} 96.47\%} & {\color[HTML]{808080} 98.07\%} & \multicolumn{1}{c|}{{\color[HTML]{808080} 94.40\%}} & {\color[HTML]{808080} 93.68\%} & \multicolumn{1}{c}{{\color[HTML]{808080} 81.24\%}} & \multicolumn{1}{c}{{\color[HTML]{808080} }} \\ \midrule
IDM                & 79.31\%                        & 86.16\%                        & 90.92\%                        & 97.33\%                        & 94.04\%                        & \multicolumn{1}{c|}{83.49\%}                        & 79.31\%                        & \multicolumn{1}{c}{79.31\%}                                                                     \\
RasterModel        & 81.64\%                        & 80.60\%                        & 86.97\%                        & 98.03\%                        & 85.04\%                        & \multicolumn{1}{c|}{81.46\%}                        & 66.92\%                        & \multicolumn{1}{c}{64.66\%}                                                                     \\
UrbanDriver        & \textbf{100.00\%}                       & 80.83\%                        & 94.13\%                        & 91.58\%                        & 90.83\%                        & \multicolumn{1}{c|}{80.28\%}                        & 67.72\%                        & \multicolumn{1}{c}{64.87\%}                                                                     \\
GameFormer         & 93.39\%                        & 89.04\%                        & 94.32\%                        & 98.67                          & 94.87\%                        & \multicolumn{1}{c|}{86.77\%}                        & 82.95\%                        & \multicolumn{1}{c}{83.88\%}                                                                     \\
PLUTO-m6           & 95.14\%                        & 87.13\%                        & 94.45\%                        & 97.87\%                        & 98.90\%                        & \multicolumn{1}{c|}{90.92\%}                        & 85.81\%                        & 76.45\%                                                                                           \\ \midrule
PlanScope-Ih10-DWT         & 95.96\%                        & 86.04\%                        & 95.37\%                        & 98.20\%                        & 98.44\%                        & \multicolumn{1}{c|}{91.74\%}                        & 85.99\%                        & 75.81\%                                                                                                \\
% PlanScope-Mh10-DWH     & 96.42\%                        & 88.72\%                        & 95.73\%                        & 98.17\%                        & 98.72\%                        & \multicolumn{1}{c|}{92.66\%}                        & 87.77\%                        & 75.39\%                                                                                                \\
PlanScope-Mh20-DWT     & 97.89\%                        & 86.70\%                        & 93.72\%                        & \textbf{98.42\%}                        & 98.81\%                        & \multicolumn{1}{c|}{91.28\%}                        & 85.80\%                        & 77.60\%                                                                                                \\ \midrule
PlanScope-Th20         & 99.27\%                        & 76.32\%                        & 95.50\%                        & 99.22\%                        & 99.36\%                        & \multicolumn{1}{c|}{91.93\%}                        & 83.86\%                        & 83.74\%                                                                                                \\
PlanScope-timedecay        & 96.24\%                        & 86.34\%                        & 95.14\%                        & 98.24\%                        & 99.63\%                        & \multicolumn{1}{c|}{91.56\%}                        & 86.48\%                        & 75.22\%                                                                                                \\
% Scope-timenorm-20\%    & 96.79\%                        & 89.57\%                        & 95.69\%                        & 98.06\%                        & 99.08\%                        & \multicolumn{1}{c|}{92.29\%}                        & 87.86\%                        &                                                                                                 \\
PlanScope-timenorm         & 98.26\%                        & 88.04\%                        & 96.28\%                        & 98.20\%                        & 99.45\%                        & \multicolumn{1}{c|}{92.94\%}                        & 88.41\%                        & 75.56\%                                                                                                \\ \midrule
PLUTO-m12-C        & 96.41\%                        & \textbf{93.28\%}                        & 96.18\%                        & 98.13\%                        & 98.53\%                        & \multicolumn{1}{c|}{93.28\%}                        & 89.04\%                        & 80.01\%                                                                                                \\
% PlanScope-Th20-m12-C        & 98.90\%                        & 87.90\%                        & 97.25\%                        & 98.33\%                        & 99.27\%                        & \multicolumn{1}{c|}{\textbf{94.59\%}}                        & 90.60\%                        & 80.14\%                                                                                                \\
PlanScope-timenorm-m12-C   & 97.25\%                        & 89.54\%                        & \textbf{97.29\%}                        & 98.04\%                        & \textbf{99.72\%}                        & \multicolumn{1}{c|}{\textbf{93.85\%}}                        & \textbf{91.32\% }                       & \textbf{80.96\%}                                                                                                \\ \midrule
\color[HTML]{808080}
PLUTO-m12-C-H      & \color[HTML]{808080} 91.93\%                        & \color[HTML]{808080} \textbf{93.65\%}                        & \color[HTML]{808080} 98.30\%                        & \color[HTML]{808080} 98.20\%                        & \color[HTML]{808080} 99.72\%                        & \multicolumn{1}{c|}{\color[HTML]{808080} 94.04\%}                        & \color[HTML]{808080} 93.21\%                        &  \color[HTML]{808080} \textbf{92.06\%}                                                                                                \\
\color[HTML]{808080}
PlanScope-timenorm-m12-C-H & \color[HTML]{808080} 93.85\%                        & \color[HTML]{808080} 93.09\%                        & \color[HTML]{808080} \textbf{98.35\%}                        & \color[HTML]{808080} 98.22\%                        & \color[HTML]{808080} \textbf{99.82\%}                        & \multicolumn{1}{c|}{\color[HTML]{808080} \textbf{94.86\%}}                        & \color[HTML]{808080} \textbf{93.59\%}                        & \color[HTML]{808080} 91.07\%                                                                                                \\ \bottomrule
\end{tabular}
    \vspace{0.5em}
    \raggedright 
    \footnotesize{
    The suffixes \textbf{m6/m12} denote that the trajectory mode number is set to 6 or 12, respectively. \textbf{Ih10} abbreviates IDD with a DS Horizon of 10, while \textbf{Mh10} refers to MDD with a DS Horizon of 10. 
    % The notation \textbf{20\%} indicates that the model is trained on 20\% of the dataset for 35 epochs, consistent with the preliminary experiments. 
    \textbf{Th20} signifies the use of Time Truncation weights with a DS Horizon of 20. \textbf{C} denotes the application of contrastive learning. \textbf{H} represents a hybrid method combining learning and rule-based post-processing. \textbf{CLS-R} represents Reactive Closed Loop Simulation.
    }
    \vspace{-2em}
\end{table*}

\subsection{Implementation Details}
Our training is conducted on the nuPlan dataset~\cite{nuplan}, which comprises a training set of 25,000 scenarios. 
We evaluate our model's performance against baseline models within the nuPlan challenge framework. 
In addition, we conduct an ablation study to determine the significance of each component within our model. 
The training is performed on a server equipped with 8 NVIDIA L20 GPUs. 
We employ the Adam optimizer with a starting learning rate of 0.001, 3 warm-up epochs, and a batch size of 32 per GPU node over 25 training epochs. 
The PyTorch library is used to implement the model, with each training session lasting approximately 2 days.

Our evaluation is conducted on the \textbf{Val14}~\cite{pdm} subset of the nuPlan dataset, encompassing 1,090 validating scenarios. Each simulation entails a 15-second rollout at a frequency of 10 Hz. 
The evaluation score is derived using the official evaluation script provided by the nuPlan challenge. Our primary comparison metric is the CLS-NR driving score, which is more challenging as other vehicles do not avoid collisions with the ego vehicle.
Besides, the future step of our model is 80 while the history step is 21. 
Both our model and \textbf{PLUTO}~\cite{pluto} are set to generate six modal trajectories.
In our experiments, the maximum horizons of decision scope across levels are controlled by setting \textbf{DS Horizon}, denoted $h$.
%\xr{Inference time is the same as PLUTO. We did not introduce any extra time consuming operation.}
Furthermore, we borrow the idea from~\cite {ptwt} to facilitate the rapid implementation and execution of wavelet transformations in our experiments.

% Thanks to Moritz \textit{et al.}, their work~\cite{ptwt} facilitates the rapid implementation and execution of wavelet transformation in our experiments
% \vspace{-1em}
\subsection{\xr{Preliminary Experiments}}

% 为了证明我们提出的MultiScope Loss的有效性，我们进行了消融实验，将消融实验的结果放在了表2中
% 在第一个实验中我们在20%的数据集上训练了baseline模型，没有使用直接速度损失，递归解码器和horizon
% 在第二个实验中我们加入了递归解码器，用于证明递归解码器的有效性
% 在第三个实验中我们将位置作为MSL的对象，用于证明我们的MSL是有效的。
% 可以看到，使用位置作为MSL的对象可以达到85.80\%，比baseline模型高了4.33%。
% 由于位置信息几乎不具备任何周期性， 且具有累加性，我们猜测其可能无法被有效分解，因此我们考虑使用速度作为MSL的对象。
% 在第二个实验中我们加入了直接速度损失，用于证明我们的scope loss是有效的，而不是直接通过提高对velocity loss的权重来提高性能
% 在接下来使用IDD且以速度作为MSL的对象的实验中，我们发现其score可以达到86.77\%，比baseline模型高了5.30%，比直接速度损失高了3.17%。
% 在后续的实验中，我们尝试了不同的horizon，用于寻找递归解码器在什么horizon更有效
% 由于我们一共有101个时间步，因此50个时间步就是0level detail的horizon长度，即使用所有的数据用于训练
% 从实验结果中我们可以看出，递归解码器在horizon为20时性能最好
% 并且可以看出，当我们使用horizon为50时，性能下降了3.87%。
% To demonstrate the effectiveness of decision scope mechanism and IDD, ablation studies are conducted. 
\hj{To identify truly effective strategies from the ones we explored, extensive comparative experiments are conducted.}
In these studies, all models are trained on 20\% of the dataset for 35 epochs and evaluated on the \textbf{Random14}~\cite{pdm} test set, comprising 249 scenarios.
The results are shown in Table~\ref{tab:ablation}. 
The ablation study encompasses the following configurations:
\xr{In the first experiment, we train the baseline model without additional peripheral losses, achieving a CLS-NR score of 83.28\%.
Rows 1-4 are designed to explore the existence of the PlanScope problem.
In these experiments, we apply a Time Truncation Horizon of 10, 20, 40, and 80 time steps, respectively. The corresponding results are 80.12\%, 87.92\%, 85.22\%, and 83.28\%.
The initially rising and subsequently falling score suggests that a planning horizon of 20 is relatively optimal. This corroborates our hypothesis that short-term decisions can improve planning performance.
However, overly myopic decisions may introduce other issues, such as collisions due to the absence of predictive planning.
Results from rows 5-6 demonstrate that weighting the regression loss enhances planning performance in closed-loop simulations.
Specifically, the Decreasing Certainty weight is denoted as \textit{timedecay}, and Time-Dependent Normalization is denoted as \textit{timenorm}.
The demonstrated value of $p=1$ and $l=e$ for \textit{timedecay} achieves the best performance in our preliminary grid search.
% Row 7 shows \textit{timenorm} is of good compatibility with the contrastive loss, which is firstly introduced in~\cite{pluto}. 
% Comparisons in rows 7–10 indicate that the Multi-head Detail Decoder exhibits better compatibility with DWH than IDD, while IDD performs well in learning supervision signals obtained via DWT. Additionally, the comparison of rows 9–12 shows that the combination of IDD and DWT achieves relatively better performance when the DS Horizon is set to 20.
Comparisons in rows 7-10 indicate that IDD with DS Horizon 10 performs well in learning supervision signals obtained via DWT.
}
\subsection{Comparison with Baselines}


% 我们将我们的算法与现有的有较好效果的方法进行了对比，并且将对比结果放在了表1中
We conduct a comparative analysis of our method against several standard approaches; the comparative results are presented in Table~\ref{tab:sota}. 
\textbf{Log-Replay}, which entails a straightforward replay of expert trajectories, serves as a baseline reference. 
This demonstrates the capability of our simulator to assess the performance of planning algorithms accurately. 
\textbf{IDM}~\cite{IDM} employs the Intelligent Driver Model to maintain safe distances from other agents and adhere to the reference lane. 
\textbf{RasterModel}, introduced in~\cite{nuplan}, is a CNN-based planning model. 
\textbf{UrbanDriver}~\cite{urban} is a vectorized planning model leveraging PointNet-based polyline encoders and transformer architecture. 
\textbf{PLUTO}~\cite{pluto}, our primary baseline, achieves state-of-the-art performance in the nuPlan planning challenge in April 2024. To save computing resources, the multimodal trajectory mode number is set to m6 at first.
By default, we have excluded the rule-based trajectory selector, thereby demonstrating that our new model enhances the performance of the neural planner.
Besides, we also eliminate the contrastive learning module, which is empowered by rule-based positive and negative sample construction.
These modules are removed to reduce artifacts that may introduce bias into the results.
% 我们的模型marked as Scope则分别测试了max horizon为10与20的情况
% 从表格中可以看出，当horizon为20时，总分最好，达到了85.97%。
% 并且在关键的安全性指标，如Time to Collision，Ego-at-fault Collision等方面，性能有较大幅度的提升。
% 而当horizon为10时，总分虽略有下降，但在几乎所有的指标上都超过了baseline。
% 值得注意的是，我们的实验并未针对性得对训练参数进行优化，因此我们相信我们的模型在进一步调优后，性能还有较大的提升空间。
Our model, designated as \textbf{PlanScope}, is evaluated with various modules that perform well in preliminary experiments. 
Part of qualitative comparison is depicted and introduced in Fig.~\ref{fig:qualitative}.
% The results, as depicted in the table, indicate that the optimal total score of 85.97\% is achieved at a horizon of 20. Notably, key safety metrics, including Time To Collision~(\textbf{TTC}) and the absence of Ego-at-fault Collisions~(\textbf{Collisions}), exhibit significant enhancements of 3.02\% and 0.96\%, respectively. 
% At a horizon of 10, although the total score is marginally reduced, our model still surpasses the baseline across nearly all metrics, encompassing Progress Along Expert (\textbf{Progress}), Collisions, Speed Limit Compliance~(\textbf{Speed Limit}), Drivable Area Compliance~(\textbf{Drivable}), and TTC\@.}
The results in the table indicate that incorporating \textit{timenorm} further improves the driving score in closed-loop simulations, achieving a total score of 93.59\%. The capabilities of the post-processing module primarily constrain this improvement. 
Under a framework that relies solely on network-based output without post-processing, our closed-loop total score 91.32\% increases by 2.28\% compared to the previous result.
In experiments with mode number equals 6 and without contrastive learning, the model with \textit{timenorm} also achieves the best CLS-NR score of 88.41\%, which increases 2.6\% compared to PLUTO-m6.
We also observe that the score for \textit{Time Truncation} is lower than the baseline, suggesting that a decision-making strategy focused solely on the short term may overlook critical long-term aspects.
Additionally, methods based on detail generation and decomposition show slight improvement.
% This may be attributed to the predefined detail layers and DS Horizons not fitting all data optimally.
More decomposition methods, decomposing profiles, wavelet forms, and related research on horizon adaptation might be required to enhance its performance further.
% \sout{Furthermore, we also found that the CLS-R score improved with }\textit{Time Truncation}. 
% \sout{The specific mechanism that lead to improvements on CLS-R score needs to be further explored.}

% More research and adaptation is required.
% We speculate learning on the decomposed trajectories provides stronger generalization ability.
% However, the CLS-R Score improves significantly. 
% Finally, the \textit{timenorm} strategy significantly improves the convergence and generalization of the neural network. 
% Notably, our method trained on only 20\% of the dataset achieves a score of 87.86\%, which is remarkably close to the 88.41\% obtained by training on the full dataset.

% It should be noted that our experiments did not involve the optimization of training parameters, suggesting that there is considerable potential for performance improvement in our model with further optimization.

% Our model attained a final score of 89.61\%, marking a 0.81\% improvement over the next best-performing method, Pluto. 
% Additionally, our model demonstrated optimal results concerning the absence of ego-at-fault collisions and compliance to speed limits. 
% These findings underscore the efficacy of our model in producing trajectories that are both more safe and more efficient.

\section{Conclusions}

This paper identifies the reasoning gap of imitation learning for urban autonomous driving  and formulates it as PlanScope, which has been overlooked in previous studies. 
\TODO{The proposed WT-based trajectory decomposition and analysis technique, DD structures, and MSS strategies jointly enable trajectory refinement across decision scopes. 
These design eliminates unpredictable future events and their negative effects for training learning-based planners.
Experiments on the nuPlan benchmark confirm that explicitly modeling motion scope yields consistent improvements in safety and efficiency metrics. 
% Future work will explore adaptive wavelet selection and joint perception-planning co-learning to improve performance further.
}
% \sout{Several approaches for enabling the model to extract necessary decisions from log replays are proposed and explored.}
Among these methods, the implementation of \textit{timenorm} achieves a state-of-the-art 91.32\% CLS-NR score in no-post-processing mode. In hybrid mode, it achieves a 93.59\% CLS-NR score, which significantly surpasses the baseline.
In addition, detail generation modes combined with our proposed decomposition methods show limited improvement. 
Their performance does not surpass that of \textit{timenorm} and \textit{timedecay}. 
Nonetheless, wavelet transformation provides a potential solution to this issue. 
We present these findings as a starting point for narrowing the exploration range for future research.
In the future, we plan to conduct more thorough experiments to explore the mechanisms underlying this problem, including further investigation of combinations and loss balancing across levels of detail.

% Future work
% 尝试更多其他小波，或对母小波进行定义
% 与感知信息的结合

% \section*{Acknowledgment}
% \appendices
% \section{Diagram of Wavelet Decomposition} \label{apx:details}
% \begin{figure}[H]
%   \centering
%   \includegraphics[width=\linewidth]{imgs/details.pdf}
%   \caption{The ground truth velocity is decomposed into different levels of details. Only the red cross-marked data points in each level are used for training.}
%   \label{fig:details}
% \end{figure}

\normalem
\bibliographystyle{IEEEtran}
\bibliography{ref}

\end{document}